\pdfoutput=1

\documentclass[11pt]{article}

\usepackage[preprint]{acl}

\usepackage{times}
\usepackage{latexsym}

\usepackage[T1]{fontenc}

\usepackage[utf8]{inputenc}

\usepackage{microtype}

\usepackage{inconsolata}

\usepackage{graphicx}

\usepackage{subfigure}
\usepackage{enumitem}
\usepackage{lipsum}
\usepackage{multirow}
\usepackage{graphicx}
\usepackage{amsmath}
\usepackage{capt-of}
\usepackage{booktabs}
\usepackage{varwidth}
\usepackage{wrapfig}
\usepackage{xspace}
\usepackage{float}
\newsavebox\tmpbox

\usepackage{amsfonts}

\newcommand{\our}{{\textsc{OmniInput}}\xspace}
\newcommand{\smallsection}[1]{\noindent\textbf{#1.}}

\title{Evaluation of human-model prediction difference \\ on the Internet Scale of Data}

\author{
 \textbf{Weitang Liu\textsuperscript{1}},
 \textbf{Ying Wai Li\textsuperscript{2}},
 \textbf{Yuelei Li \textsuperscript{1}},
 \textbf{Zihan Wang\textsuperscript{1}},
 \textbf{Yi-zhuang You\textsuperscript{1}}
 \textbf{Jingbo Shang\textsuperscript{1}}
\\
\\
 \textsuperscript{1}University of California, San Diego \\ La Jolla, CA 92093, USA \\
 \textsuperscript{2}Los Alamos National Laboratory \\ Los Alamos, NM 87545, USA
\\
 }

\begin{document}
\maketitle

\begin{abstract}
Evaluating models on datasets often fails to capture their behavior when faced with unexpected and diverse types of inputs.
It would be beneficial if we could evaluate the difference between human annotation and model prediction for an internet number of inputs, or more generally, for an input space that enumeration is computationally impractical. 
Traditional model evaluation methods rely on precision and recall (PR) as metrics, which are typically estimated by comparing human annotations with model predictions on a specific dataset. This is feasible because enumerating thousands of test inputs is manageable.
However, estimating PR across a large input space is challenging because enumeration becomes computationally infeasible.
We propose \our, a novel approach to evaluate and compare NNs by the PR of an input space. 
\our is distinctive from previous works as its estimated PR reflects the estimation of the differences between human annotation and model prediction in the input space which is usually too huge to be enumerated.
We empirically validate our method within an enumerable input space, and our experiments demonstrate that \our can effectively estimate and compare precision and recall for (large) language models within a broad input space that is not enumerable.

\end{abstract}

\section{Introduction~\label{sec:intro}}

\begin{figure*}[t]
  \centering
\includegraphics[width=\linewidth]{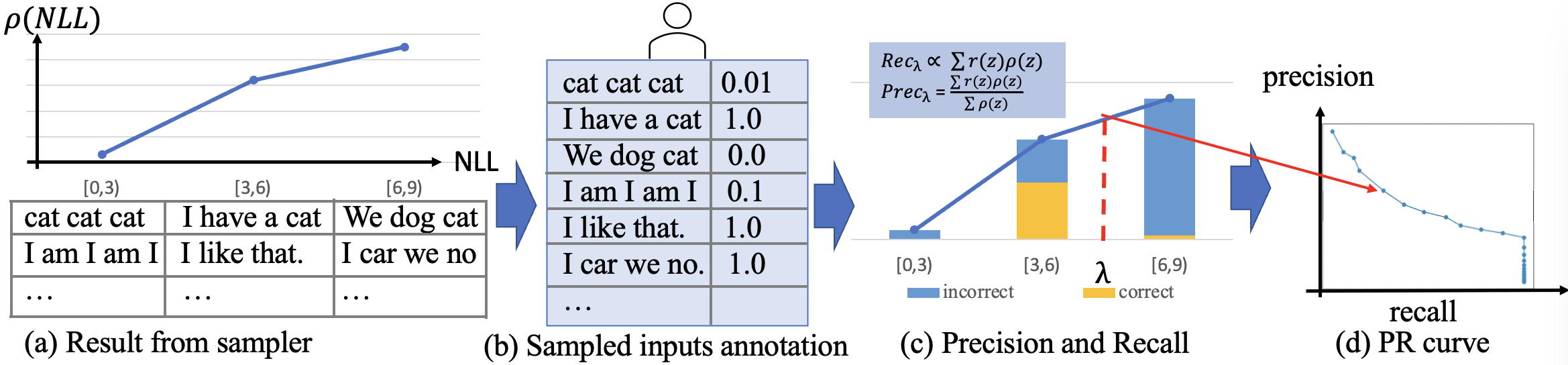}
  \vspace{-4mm}
  \caption{
  An overview of our novel \our framework. 
  (a) Use an efficient sampler to obtain the output distribution $\rho(z)$ and the sampled inputs;
  (b) Annotate the sampled inputs; 
  (c) Estimate the precision and recall at different threshold $\lambda$ that distinguishes different classes.
  $r(z)$ denotes the precision of the model within the bin of output value $z$; 
  (d) Construct a precision-recall curve.
  }
  \label{fig:intro}
  \vspace{-5mm}
\end{figure*}

Recently, neural network-based agents have been deployed online, making them widely accessible to the public. 
his presents a challenge, as users can input virtually any type of data, potentially causing the models to behave unpredictably. 
It is common to collect a dataset of specific types of misbehavior and test models on this dataset to assess the misbehavior, where precision and recall (PR) are the commonly used metrics to assess the model performance with thousands of data points~\citep{liu2020energy,hendrycks2016baseline,hsu2020generalized,lee2018simple, szegedy2013intriguing}. 
However, these datasets with a limited number of inputs often fail to accurately represent the behaviors of models that are likely to encounter a vast number of inputs when deployed online. 
It would be beneficial if we could assess the model performance of human-model prediction difference on the internet number of inputs, or more generally, on a (discrete) input space which is finite but computationally impossible to enumerate all the data points.

Assessing model behaviors using datasets, such as for privacy leaks, often fails to capture the full range of unforeseen behaviors that could emerge from the internet number of inputs encountered online.
A simple approach is to uniformly sample data from the internet, 
feed these inputs to the models, and compare the models' predictions with human annotations to generate PR metrics.
However, this uniform sampling strategy is generally impractical for the vast number of potential inputs online because models typically do not predict with high confidence on most sampled inputs. They are confident with the inputs that they believe they are familiar with, such as those from their training distribution, or with overconfident inputs.

We propose \our that leverages the output distribution to obtain the precision-recall of the input space of innumerable inputs as if we were uniformly sampling the input space. The output distribution is the count of the inputs that correspond to each output value, showing the relative difference in the number of inputs for different outputs. This quantity is the key to estimating the precision and recall in an input space where enumeration is impossible. As shown in Fig.~\ref{fig:intro}, it consists of four steps:
\begin{enumerate}[nosep,leftmargin=*]
    \item[(a)]
        We employ a recently proposed sampler to obtain the output distribution $\rho(z)$ of the trained model (where $z$ denotes the output value of the model) over an input space~\citep{pmlr-v202-liu23aw} and efficiently sample the inputs from different output value (e.g., negative-log-likelihood) bins.
        The output distribution is a histogram counting the number of inputs that lead to the same model output. 
        In the open-world setting without any prior knowledge of the samples, all possible inputs should appear equally. 
    \item[(b)] 
        We annotate the sampled inputs, e.g., rate how likely the inputs are understandable sentences using a score from 0 to 1 for language models.
    \item[(c)]
        We compute the precision for each bin as $r(z)$, then estimate the precision and recall at different threshold values $\lambda$. 
        When aggregating the precision across different bins, a weighted average of $r(z)$ by the output distribution $\rho(z)$ is required i.e., $\frac{\sum_{z \le \lambda} r(z)\cdot \rho(z)}{\sum_{z \le \lambda}{\rho(z)}}$. See Sec.~\ref{model-centric} for details.
    \item[(d)]
        We finally put together the precision-recall curve for a comprehensive evaluation of the model performance over the input space. 
\end{enumerate}

\our samples the inputs solely by the model itself, eliminating possible human biases introduced by the test data collection process~\cite{ luo2023zeroshot,prabhu2023lance,shu2020identifying,leclerc20223db}.
The resulting precision-recall curve can help decide the limit of the model in real-world deployment. 
A model with a high area under the precision-recall (AUPR) curve in \our is expected to agree closely with human's annotations.

Our experiments using \our reveal, for the first time, the prediction differences between humans and models through the precision-recall curves in a large input space that is not enumerable.
We apply \our to (large) language models (LLMs), GPT2~\cite{radford2019language}, Llama2~\cite{touvron2023llama}, and DistilBERT (for text sentiment classification).
Through these newly estimated precision-recall curves, \our makes it possible to compare the models beyond datasets to an input space. 
Our experiments do not serve as a conclusive study of the models trained with different training methods and architectures.
Instead, we view \our as a proof of concept, paving the way for future research that seeks to better understand the prediction differences between humans and models in the vast, non-enumerable input space. Our new contributions are:
\begin{itemize}[nosep,leftmargin=*]
    \item We propose to understand an AI/ML model's input-output mapping solely by utilizing the model itself beyond the pre-defined datasets -- the input space.   
    \item We develop a novel model understanding framework, \our, for humans to inspect the inputs and to compute the precision and recall in the input space by leveraging output distribution, enabling the understanding of the model's input-output mapping in input space. 
    \item We apply \our to evaluate various popular (large) language models. The results reveal the first time the precision and recall of different models between human annotations and model predictions in a large and non-enumerable input space.
\end{itemize}

\section{The \our Framework}
In this section, we present a detailed background on sampling the output distribution across the input space. 
We then propose a novel framework \our for humans to understand the precision and recall in the input space. 

\subsection{Output Distribution and Sampler}

\smallsection{Text Generation by Sampling} 
Generating text by sampling is popular in natual language processing~\cite{kumar2022constrained, qin2022cold}. They use Markov Chain Monte-Carlo (MCMC) and the input space is not assumed to be enumerable. 
As pointed out~\cite{du2023principled}, text generation should employ samplers of discrete input space~\cite{goshvadi2024discs, grathwohl2021oops, zhang2022langevinlike}.
The target distribution for all these samplers is 
\begin{equation}~\label{equ:mcmc}
    p(\mathbf{x}) \propto \exp (g(\mathbf{x})/T),
\end{equation}
where $g(\cdot)$ is the (negative) ``energy'' and $T$ is a temperature. When $T$ is 1, $g(\cdot)$ becomes the log-probability, a common quantity to be modeled in machine learning~\cite{lecun2006tutorial}. 
This $p(\mathbf{x})$ in Equ.~\ref{equ:mcmc} is also a widely used target distribution in machine learning sampler community. 

\our uses the exact same sampling setting of discrete inputs and this is the major bottleneck of Model-diff. The time-complexity of Model-diff is therefore similar to text generation by sampling. 
Post-processing of Model-diff after text generation by sampling only takes few hours.

\smallsection{Output Distribution}
We denote a trained binary neural classifier 
parameterized by $\theta$ as $f_\mathbf{\theta}: 
\mathbf{x} \rightarrow z$ where $\mathbf{x} \in \Omega_T$ is the training set, 
$\Omega_T \subseteq \{0,...,N\}^D$, and $z \in 
\mathbb{R}$ is the output of the model. 
In our framework, $z$ represents the logit and each of the $D$ pixels takes one of the $N+1$ values. The output distribution represents the frequency count of each output logit $z$ given the  (entire) input space $\Omega = \{0,...,N\}^D$ or some other space $\Omega=\Omega_M$ specified by a model M. 
In our framework, following the principle of equal \emph{a priori} probabilities, 
we assume that each input sample within $\Omega$ follows a uniform distribution. This assumption is based on the notion that every sample in the input space holds equal importance. Mathematically, the output distribution, denoted by $\rho(z)$, is defined as:
\[
\rho(z) = \sum_{\mathbf{x} \in \Omega} \delta(z-f_\mathbf{\theta}(\mathbf{x})),
\]
where $\delta(\cdot)$ is $1$ if the input is $0$ or is $0$ otherwise.

\smallsection{Samplers} The sampling of an output distribution finds its roots in physics, particularly in the context of the sampling of the density of states (DOS)~\citep{wang2001efficient,vogel2013generic, cunha2008improving, junghans2014molecular,li2017histogram, PhysRevLett.96.120201}, but its connection to ML is revealed only recently~\citep{pmlr-v202-liu23aw}. 
The output distribution $\rho(z)$ is unknown in advance.   
In practical implementations, the ``entropy'' (of discretized bins of $z$), $\tilde S(z) = \log \tilde\rho(z)$, is used to store the instantaneous estimation of the ground truth $S(z) = \log \rho(z)$. 

Parallel Tempering and Histogram Reweighting (PTHR)~\cite{hukushima1996exchange, swendsen1986replica} is an efficient approach to sample output distribution. 
It starts with the same target distribution for Markov chain Monte-Carlo (MCMC) sampler~\cite{grathwohl2021oops, zhang2022langevinlike}:
\begin{equation}~\label{equ:mcmc}
    p(\mathbf{x}) \propto \exp (f_\theta(\mathbf{x})),
\end{equation}
where the model output $f_\theta(\cdot)$ is also called (negative) ``energy'' (log-probability), if it learns to model log-probability~\cite{lecun2006tutorial}. 
After the MCMC sampling, PTHR reweights the sampled distributions to acquire the output distribution, because the MCMC samplers sample more often on $\mathbf{x}$ whose output $f_\theta(\mathbf{x})$ is larger. 
PTHR is compatible with MCMC samplers and therefore it can take advantage of the development of MCMC samplers that follow the same target distribution Equ.~\ref{equ:mcmc}.

\subsection{Precision-Recall of the Input Space}
\label{model-centric}
\our revolves around the \emph{output distribution} to formulate the \emph{estimation} of the precision-recall from evaluated sampled inputs to the input space. 

\smallsection{Generating Text as Inputs by Sampling} Sampling methods are common in text generation in language models~\cite{goshvadi2024discs, kumar2022constrained, qin2022cold}. As pointed out~\cite{du2023principled}, text generation should use the samplers for discrete input space, which \our adopts. As \our generates text using the sampling method, the cost is similar to text generation using sampling.


\smallsection{Annotation of Inputs} Our motivation is for humans to understand the prediction difference between humans and models and therefore humans serve as a gold standard in annotation.
After humans understand the training distribution by scrutinizing the training set, they annotate a score to each input within the same ``bin'' of the output distribution (each ``bin'' collects the inputs with a small range of output values $[z-\Delta z, z+\Delta z)$). 
This score ranges from $0$ when the sample completely deviates from the annotator's judgment for the target class, to $1$ when the prediction of the input perfectly agrees with the annotator's judgment (``\textbf{good}'' input). Following the evaluation, the average score for each bin, termed ``precision per bin'', $r(z)$, is calculated. It is the proportion of the total evaluation score on the inputs relative to the total number of inputs within that bin. We have 150-600 bins for the experiments.

\smallsection{Precision and Recall (PR)~\footnote{In the input space, another commonly used metric ROC is closely connected to PR (Appendix~\ref{app:ROC}).}}
For our experiments of language models, we use the negative-log-likelihood (NLL~\footnote{NLL is the log perplexity.}) as the output $z$ for sampling, because it is the loss of next-token prediction, but other quantities can also be used for sampling for specific tasks. 
We define a varying threshold of model confidence $\lambda$ such that any inputs predicted with $z \leq \lambda$ by the model are from the training distribution. 
Thus, the precision given $\lambda$ is defined as 
\vspace{-2mm}
\begin{equation}~\label{equ:precision}
\mathrm{precision}_{\lambda} = \frac{\sum_{z\leq \lambda} r(z)\rho(z)}{\sum_{z\leq \lambda} \rho(z)}.
\end{equation}
The numerator is the \emph{true positive} which is the estimate of the number of ``good'' inputs and the denominator is the total number of inputs predicted as positive -- the sum of true positive and false positive. This denominator can be interpreted as the \emph{area under curve} (AUC) of the output distribution from the $-\infty$ to threshold $\lambda$. A higher precision indicates a higher proportion of the inputs agreeing with annotators' judgments for the given output values. 

When considering recall, we need to compute the total number of ground truth inputs that the annotators labeled as the target class. This total number of ground truth inputs remains constant (albeit unknown) over the input space. Hence recall is proportional to $\sum_{z\leq \lambda} r(z)\rho(z)$: 
\begin{align}
    \mathrm{recall}_{\lambda} &= \frac{\sum_{z\leq \lambda} r(z)\rho(z)}{\text{number of positive inputs}} \notag\\
&\propto 
\sum_{z\leq \lambda} r(z)\rho(z).
\end{align}

A higher recall indicates more inputs agrees with the annotators' judgments are captured by the model. As demonstrated above, the output distribution is a valuable quantity for deriving both precision and (unnormalized) recall. These metrics can be utilized for humans to understand the model's mapping by varying the threshold $\lambda$. When $\rho(z)$ differs significantly for different $z$, precision$_\lambda$ is approximated as $ r(z^*)$ where $z^* = \arg\max_{z \geq \lambda} \rho(z)$ and recall$_\lambda$ is approximately proportional to $\max_{z \geq \lambda} r(z)\rho(z)$.

\smallsection{Application: Model Comparison} One application of \our is to compare models by PR based on each model's own sampled inputs that reflect the dominant patterns. 
To facilitate a meaningful comparison of different models, $M_1(\cdot)$ and $M_2(\cdot)$ based on their sampled inputs and output distributions, it is important to normalize the output distributions of the two models. 
To achieve the comparison, we first designate an input subspace from the original input space, such as the subspace whose inputs have their corresponding outputs predicted by both models within a certain range of $Z = [z_-, z_+]$: $X = \{\mathbf{x} | M_1(\mathbf{x}) \in Z \text{ and } M_2(\mathbf{x}) \in Z \}$.
During sampling for $M_1$, we acquire a sampled (partial) unnormalized output distribution $\rho_{M_1}$ and we find $X_{M_1}$ samples are from $X$. 
During sampling for $M_2$, we acquire the output distribution $\rho_{M_2}$ and we find $X_{M_2}$ samples are from $X$.
We can therefore get the normalized output distributions as 
\begin{align}
    \hat{\rho}_{M_1} &= \frac{\rho_{M_1}}{X_{M_1}}, \\ 
    \hat{\rho}_{M_2} &= \frac{{\rho}_{M_2}}{X_{M_2}}.
\end{align}
Both $\hat{\rho}_{M_1}$ and $\hat{\rho}_{M_2}$ are directly comparable, because $X$ is shared by both of the models.
In practice, having $Z$ is also preferable because not all output values are interesting to consider. For example, a large NLL mostly corresponds to noisy inputs and they generally have very small precision.  

As an intuitive example, suppose there is a large input space where model $M_1$ maps 500 inputs to outputs within $Z$ and Model $M_2$ maps 200 inputs within $Z$. $X$ is the set of inputs that are predicted by both models within $Z$ and $X$ has 100 inputs. 

We sample 50 inputs with outputs within $Z$ by $M_1$, and we should have around 10 of them from $X$. The sampled proportion is 50/10=5 ($\hat{\rho}_{M_1}$), which is consistent with the ground truth proportion 500/100. 
We then sample 50 inputs with outputs within $Z$ by $M_2$, and we should have around 25 of them from $X$. The sampled proportion is 50/25=2 ($\hat{\rho}_{M_2}$), which is consistent with the ground truth proportion 200/100. 

The ratio ($\hat{\rho}_{M_1}$ / $\hat{\rho}_{M_2}$) is 5/2 which is the same as the ground truth ratio 500/200. This is the ratio of the number of inputs model $M_1$ maps to $Z$ with respect to the number of inputs $M_2$ maps to $Z$. 

\smallsection{Annotation of different Sampled Inputs}
It seems unreasonable to annotate the inputs sampled by two models and compare models based on the different sets of sampled inputs, but it is a paradox. We reason with a hypothetical example. 

Suppose we pre-train two models to identify fluent sentences. After training, we present them with a vast number of sentences from the internet and ask them to select fluent ones with 99\% confidence. It's impractical to label all the sentences.
Instead, we rely on the models to first identify fluent sentences, as they can process inputs much faster than we can. The models review the sentences for an extended period (sampling), then select a few for us to annotate: one model selects sentences that simply repeat words, while the other identifies three fluent sentences and two other sentences containing noisy words. Most importantly, both models are confident they have chosen fluent sentences, but we know they make mistakes.

Is it unfair to compare the models simply because they select different inputs? Not necessarily. Despite the differences in input selection, we can objectively assess which model identifies more fluent sentences. That both models were trained with the same objective and were tasked with selecting fluent sentences from the same input space ensures a valid basis for comparison. 

It is important to recognize that the differences in the sentences each model selects directly reflect their distinct abilities, or beliefs, regarding what constitutes a fluent sentence. These differences are precisely what \our aims to compare. \our follows a similar approach, but it additionally requires varying levels of certainty to compute precision and recall, which necessitates estimating the output distribution.

\section{Experiments}

\subsection{Experimental settings}
We first apply \our to a \textbf{Toy} example where enumeration of all inputs is affordable to confirm \our's correctness (Sec.~\ref{sec:toy}). 
In Sec.~\ref{sec:real_world}, \our is used in two pre-trained GPT2 models~\cite{radford2019language} and Llama models~\cite{touvron2023llama,touvron2023llama2} with sequence length 25. 
We apply \our in longer sequences with length 100 for GPT2 models. Finally, we also apply \our to a sentiment classifier. 
The sampling target is the training loss used for next-token predictions, the negative-log-likelihood (NLL) which is also our sampling target. 

\subsection{Toy example}~\label{sec:toy}
\begin{figure}[t]
  \centering
\includegraphics[width=\linewidth]{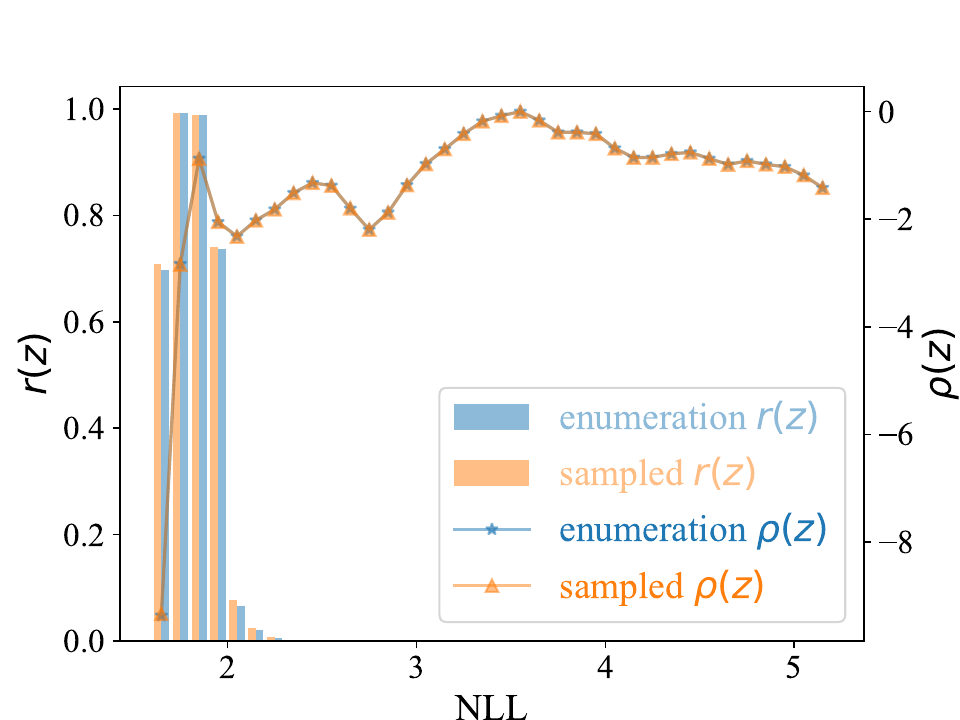}
  \vspace{-4mm}
  \caption{Toy example where enumeration is affordable. The bar plots compare the ground truth and the sampled precision per bins $r(z)$. The line plots compare the ground truth and the sampled output distributions $\rho(z)$.}
  \label{fig:toy}
  \vspace{-5mm}
\end{figure}

\textbf{Toy} is a simple GPT2 model we train to generate sequences of length 8 where the modulo of the sum of the 8 tokens by 30 is 0: $(\sum \mathbf{x})$ \text{mod} $30 = 0$. Each token is an integer \{0,1,...,9\}. Therefore, the input space is $10^8$ which is enumerable. 
The model has 4 heads and 6 layers.
We confirm the model can generate sequences where the modulo of the sum by 30 is 0 by 100\%. 
Fig~\ref{fig:toy} shows the results of $r(z)$ and $\rho(z)$ compared to the ground truth enumeration. We can confirm the sampled results are very close to the ground truth enumeration.
\textbf{Toy} validates the accuracy of the sampling results, demonstrating that they effectively represent the enumeration. This gives us confidence to apply the approach to more complex applications.

\subsection{Real-world (large) language models}~\label{sec:real_world}
We apply \our to a pretrained GPT2 and Llama models.
We apply \our to two pretrained GPT2 with 25 tokens for GPT2-medium as \textbf{GPT2-medium-25} and GPT2-small as \textbf{GPT2-small-25}. 
We also apply \our to these two models with 100 tokens for GPT2-medium as \textbf{GPT2-medium-100} and GPT2-small as \textbf{GPT2-small-100} to test on longer sequences.
To test on larger models, we apply \our to two pretrained Llama models with 25 tokens for Llama1-7B as \textbf{Llama1-25} and Llama2-7B as \textbf{Llama2-25}. We use the default vocabulary size (32,000 for Llama and 50,257 for GPT2). In the real-world setting, we remain agnostic of the training set because we apply \our to pretrained models, and we do not need test set because \our does not need predefined dataset for evaluation. 

For sampling goal NLL, the low NLL indicates the model (strongly) believes that the sequence is similar to the training distribution (e.g. fluent sentences). However, it was found very low NLL sequences contain repeating words and are difficult to be understood by humans~\cite{holtzman2019curious}.
Therefore, we preset an output range and only consider inputs whose outputs (NLL) are within the range for precision and recall (PR). Other NLL values could be used depending on the task.
In the GPT2-small-25 experiment, the model repeats the words when NLL is less than 2 and the sentences are hard to understand when NLL is larger than 5. Because of this and the limited human resources for annotation, we designate a range of outputs bins for precision and recall with at least 30 sampled inputs without duplicates and annotate 30 inputs per bin. We label the inputs with NLL ranging from $2.0$ to $4.0$ for GPT2-medium-25 and GPT2-small-25. For GPT2-small-100 and GPT2-medium-100, we annotate the outputs with NLL ranging from $4.0$ to $5.0$. For Llama2-25 and Llama1-25, NLL ranges from $3.5$ to $4.5$. Each bin captures $\Delta$NLL $=0.1$. Other NLL values can be labeled similarly.

\begin{figure*}
\begin{center}
\subfigure[PR for GPT2 with sequence length 25.~\label{fig:gpt2_25}]{\includegraphics[width=40mm]{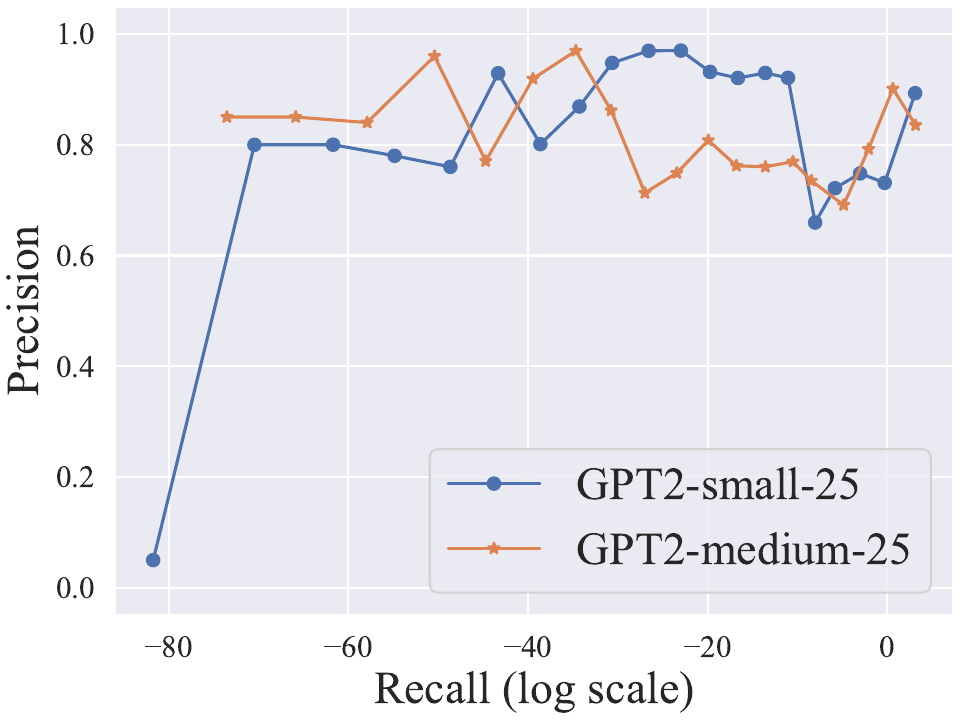}}
\subfigure[PR for GPT2 with sequence length 100.~\label{fig:gpt2_100}]{\includegraphics[width=40mm]{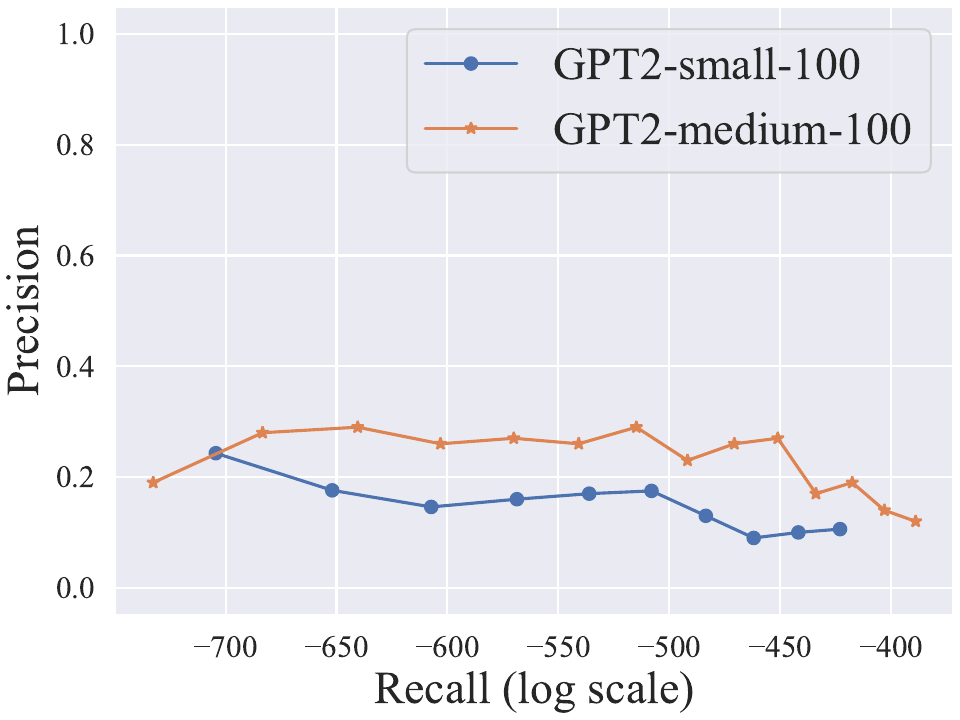}}
\subfigure[PR for Llama models with sequence length 25.~\label{fig:llama_25}]{\includegraphics[width=40mm]{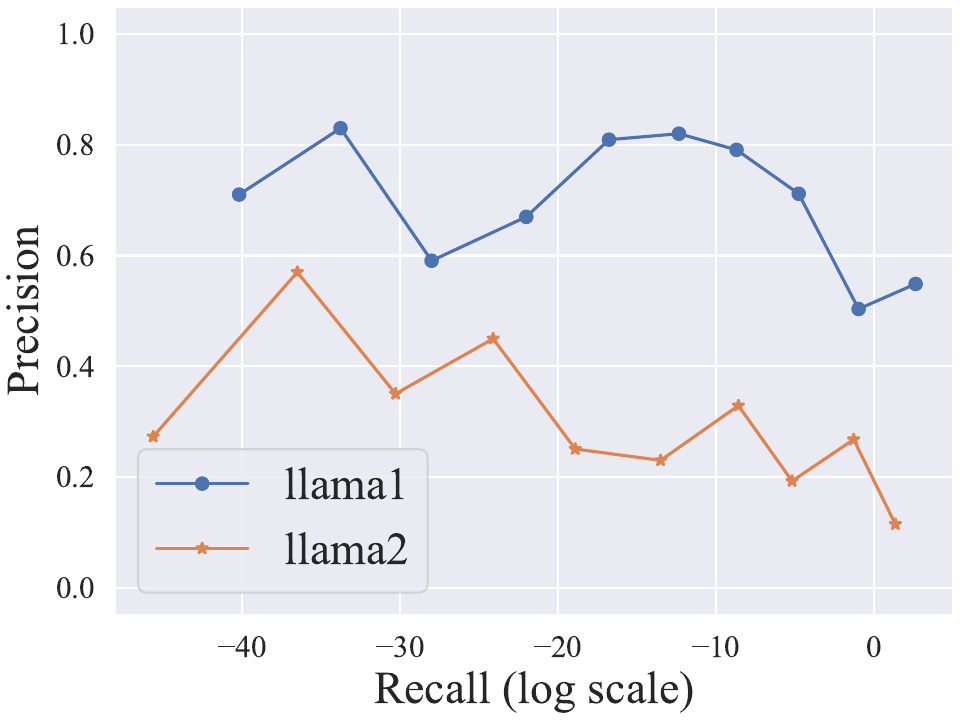}}
\caption{PR for Language models.~\label{lm}}
\end{center}
\end{figure*}

\smallsection{Results} Fig.~\ref{lm} shows \our can be applied to produce the PR curves for different models and sequence length. 
The PR curve for both GPT2 models with sequence length 25 (Fig.~\ref{lm}(a) ) shows this setting generally leads to highly understandable sequences because of the high precision. Both curves has wins and loses, indicating that they achieve similar precision and recall for sequence length 25.

The PR curve for both GPT2 models with sequence length 100 (Fig.~\ref{lm}(b) ) shows this setting generally leads to sequences that are difficult to be understood, because of the low precision. 
The PR curve for GPT2-medium-100 is almost always above the PR curve for GPT2-small-100. Integrating the area under curve of PR (AUPR) for both models respectively,  GPT2-medium has larger AUPR and thus performs better than GPT2-small for sequence length 100, though both of them achieve low precision in general.

The PR curve for both Llama models with sequence length 25 (Fig.~\ref{lm}(c) ) shows the sequences from Llama1 are easier to be understood, as the precision is higher. 
Since both models exhibit a similar range of recall, integrating the PR curve results in a higher AUPR, indicating that Llama1 outperforms Llama2 for sequence lengths of 25 and within the selected output range $Z$.

Fine-grained analysis by scrutinizing the inputs raises some concerns about privacy leaking and hallucination. For example, an input we encounter is some company names with their addresses. Our search results on these names or addresses seem to not correspond to each other, suggesting a potential privacy leak. Another example is a sequence of magic key words ``Good morning dear friend, I, and Greetings, ladies and gentlemen''. GPT2-small-25 keeps generating email addresses before this sequence. This raises a concern of privacy leaking of the models. 
For GPT2-small-100, we can sample NLL down to around $2.7$ where we find repeating words similar to sentences with NLL smaller than 2 in GPT2-small-25. When NLL gets higher, the repeating phrases are generally more meaningful, such as ``pickup truck pickup 4 trailer trailer'' or ``put clean clothes put things wash clothes'', compared to simply non-meaningful words in DistilBERT. Overall, sentences with 100 tokens in the selected range repeat the phrases and generally are not fluent. 
\our applied to Llama2-25 finds inputs that have extreme human annotation scores for the selected range of NLL: either (relatively) low or (relatively). More experimental details are in Appendix~\ref{app:llm}. Based on our results, we speculate that the next-token generation ability of these models, driven by a sophisticated next-token generation function, may not fully align with the NLL for an entire sentence.

\subsection{Language classifier}
We fine-tune a DistilBERT~\citep{Sanh2019DistilBERTAD} using SST2~\citep{socher-etal-2013-recursive} and achieve ~91\% accuracy. We choose the logits as our sampling target.  
We evaluate this model using \our.
Since the maximum length of the SST2 dataset is 66 tokens, one can define the input space as the sentences with exactly 66 tokens. 
For shorter sentences, the last few tokens can be simply padding tokens.
One might be more interested in shorter sentences because a typical sentence in SST2 contains 10 tokens.
Therefore, we evaluate length 66 and length 10 sentences, respectively. For sentence length equals 66, we have 15 bins with around 200 inputs per bin. 

\smallsection{Results} We apply \our to a fine-tuned DistilBERT~\citep{Sanh2019DistilBERTAD} on the SST2 dateset~\citep{socher-etal-2013-recursive}. The sampled inputs per bin for each logit reveal that the model tends to classify positive sentiment primarily based on specific positive keywords, rather than understanding the grammar and structure of the sentences.
Appendix~\ref{sup:sst} contains more sampled inputs.
For sentence length 66, some sampled inputs with logit equals 7 (positive sentiment) in Fig~\ref{fig:sst_long}. For sentence length 10, some sampled inputs with logit equals to 7 (positive sentiment) in Fig~\ref{fig:sst_short}.

\smallsection{Discussion} In the open-world, it is understandable that the language classifier performs poorly because the classifiers are trained to predict
the conditional probability $p(class|\mathbf{x})$ where $\mathbf{x}$ are from the training distribution. 
To deal with the open-world setting, the models also have to learn the data distribution $p(\mathbf{x})$ in order to tell whether the inputs are from the training distribution. 
Additionally, our method indicates an importance of recall besides precision that can be normally estimated by the proportion of the ``good'' inputs compared to the inputs predicted as positive including both true positive and false positive. In summary, our method not only estimates the precision of the models but also their recall. This dual capability provides a pathway to enhance model performance by improving both metrics across a large input space.

\smallsection{Human Annotation Ambiguity}

\begin{table}
  \centering
  \footnotesize
  \begin{tabular}{lc}
    \hline
    \textbf{NLL range} & \textbf{$r(z)$} \\
    \hline
    $[3.5, 3.6)$ & $0.19\pm0.05$\\
    $[3.6, 3.7)$ & $0.25\pm 0.04$ \\
    $[3.7, 3.8)$ & $0.26\pm0.01$\\
    $[3.8, 3.9)$ & $0.23\pm0.04$\\
    $[3.9, 4.0)$ & $0.25\pm0.01$\\
    $[4.0, 4.1)$ & $0.26\pm0.01$\\
    \hline
  \end{tabular}
  \caption{Human annotation variance estimation. }
  \label{tab:human_ambiguity}
\end{table}

We observe that different models exhibit varying degrees of inconsistencies in human labeling. 
As a demonstration study, we examine the variations in $r(z)$ when two different individuals label the same dataset (Tab.~\ref{tab:human_ambiguity}).

\section{Related Works}
\noindent \textbf{Performance Characterization} has been extensively studied in the literature~\cite{haralick1992performance, klette2000performance, thacker2008performance, ramesh1997computer, bowyer1998empirical, aghdasi1994digitization, ramesh1992random, ramesh1994methodology}. Previous research has focused on various aspects, including simple models~\cite{hammitt1995determining} and mathematical morphological operators~\cite{gao2002statistical, kanungo1990character}.
In our method, we adopt a black box setting where the analytic characterization of the input-to-output function is unknown~\citep{courtney1997algorithmic, cho1997performance}, and we place emphasis on the output distribution~\citep{greiffenhagen2001design}. This approach allows us to evaluate the model's performance without requiring detailed knowledge of its internal workings.
Recent works~\citep{qiu2019semanticadv,explaining_in_style, luo2023zeroshot,prabhu2023lance} evaluate model performance without test set. They used other generators to generate samples for evaluating a model. On the contrary, we used a sampler to sample the model to be evaluated. Sampling is transparent with convergence estimates, but other generators are still considered as black boxes. Given the inherently unknown biases in models, utilizing other models to evaluate a model carries the risk of yielding unfair and potentially incorrect conclusions. Our method brings the focus back to the model to be tested, tasking it with generating samples by itself for scrutiny, rather than relying on external agents such as human or other models to come up with testing data. An additional benefit is that this approach offers a novel framework for estimating errors in the input space when comparing different models. 
 
\vspace{-3mm}
\paragraph{Samplers} 
MCMC samplers have gained widespread popularity in the machine learning community~\citep{chen2014stochastic, welling2011bayesian, li2016preconditioned, xu2018global}. Among these, CSGLD~\citep{CSGLD} leverages the Wang--Landau algorithm~\citep{wang2001efficient} to comprehensively explore the energy landscape. Gibbs-With-Gradients (GWG)\citep{grathwohl2021oops} extends this approach to the discrete setting, while discrete Langevin proposal (DLP)\citep{zhang2022langevinlike} achieves global updates.
Although these algorithms can in principle be used to sample the output distribution, efficiently sampling it requires an \emph{unbiased} proposal distribution. As a result, these samplers may struggle to adequately explore the full range of possible output values. Furthermore, since the underlying distribution to be sampled is \emph{unknown}, iterative techniques become necessary. The Wang--Landau algorithm capitalizes on the sampling history to efficiently sample the potential output values. The Gradient Wang--Landau algorithm (GWL)~\citep{pmlr-v202-liu23aw} combines the Wang--Landau algorithm with gradient proposals, resulting in improved efficiency. 

\noindent \textbf{Open-world Model Evaluation} requires model to perform well in in-distribution test sets~\citep{dosovitskiy2020vit, tolstikhin2021mixer, steiner2021augreg, chen2021outperform,zhuang2022gsam, He2015,simonyan2014very, szegedy2015going, huang2017densely, zagoruyko2016wide}, OOD detection~\citep{liu2020energy,hendrycks2016baseline, hendrycks2018deep,hsu2020generalized,lee2017training,lee2018simple,liang2018enhancing, mohseni2020self, ren2019likelihood}, generalization~\citep{cao2022openworld, sun2022open}, and adversarial attacks~\citep{szegedy2013intriguing,rozsa2016adversarial,miyato2018virtual,kurakin2016adversarial,xie2019improving,madry2017towards}. Understanding performance of the model needs to consider the input space that includes all these types of samples.

\section{Conclusions~\label{sec:con}}
\vspace{-2mm}
We introduce \our, a new framework to help humans understand the precision and recall of an input space. 
As future work, developing efficient samplers for output distribution is crucial, yet it has received limited attention in the community.
Our work demonstrated the importance of sampling from output distribution by showing how it enables the understanding of model's input-output mapping in an input space. 

\section{Assumptions, limitations, and future work of \our~\label{app:limitation}} Our framework is designed to be a general framework, but may not be preferable for all settings.
First, we may need to be careful when humans cannot serve as a gold standard for evaluation, such as when humans have significantly different views in annotations. 
Due to limited resources and the nature of our proof-of-concept project, we estimated human annotation ambiguity and effort on a small scale. Estimating on a larger scale and addressing the elimination of human involvement will be considered in future work.

Second, our analysis depends on the sampler(s). As sampling the output distribution is a relatively new topic in the machine learning community, more advanced samplers with more computation resources can scale our experiments. Although our proof-of-concept method depends on the samplers' results, the analysis method itself is parallel to the development of the sampler, meaning that the method of how to use output distributions to analyze the models will be consistent, even though the sampled results may improve with better samplers. 

Third, \our relies on the model's own sampled inputs to understand the model itself. As the sampled inputs from one model can simply be subdominant for another model, cross-checking sampled inputs among models needs to overcome the challenge of aligning the counting of sampled inputs from different models. Because subdominant inputs do not reflect the dominant characteristics of the output value(s), we leave cross-checking among models as future work.

\clearpage
\appendix
\section{Broader Impact~\label{sec:broad}}
This paper aims to provide a comprehensive evaluation of models. However, there are important societal implications to consider. One concern is that the model might retain and potentially recover training data through sampling. Data privacy must be a key consideration when such models are presented within the ML community.

\section{Sampler Details~\label{sup:alg}}

\subsection{sampler with same $\beta$}

Gradient-with-Gibbs (GWG) is a Gibbs sampler by nature, thus it updates only one pixel at a time. Recently, a discrete Langevin proposal (DLP)~\cite{zhang2022langevinlike} is proposed to achieve global update, i.e., updating multiple pixels at a time. We adopt this sampler to traverse the input space more quickly, but we treat $-\frac{d \tilde{S}}{d f}$ the same value as $\beta$ for both $q(\mathbf{x}'|\mathbf{x})$ and $q(\mathbf{x}|\mathbf{x}')$. We sample $\beta$ uniformly from a range of positive and negative values to balance small updates ($|\beta|$ is small) and aggressive updates ($|\beta|$ is large). The forward and backward proposal probabilities will share the same $\beta$, which improves the efficiency of the sampling. 

\section{Connection between ROC and PR curve in the input space~\label{app:ROC}}
It is common in the traditional evaluation framework to consider the receiver operating characteristic curve (ROC) and precision-recall (PR) separately. The recall in PR is the same as the unnormalized true positive rate in ROC, so we do not need to consider the true positive rate separately. The false positive is the number of positively predicted inputs minus the number of the true positives (using the notation of Equ.~\ref{equ:precision})
\begin{equation*}
\sum^{+\infty}_{z\geq \lambda} \rho(z)- \sum^{+\infty}_{z\geq \lambda} r(z)\rho(z)=\sum^{+\infty}_{z\geq \lambda} \rho(z)(1-r(z))
\end{equation*}
The false positive rate is the number of false positives divided by the number of inputs of the negative class. Since the number of inputs of the negative class is a constant in the input space, the unnormalized false positive rate is:
\begin{equation*}
\text{False positive rate} \propto \sum^{+\infty}_{z\geq \lambda} \rho(z)(1-r(z)).
\end{equation*}
In other words, once we compute the true positive, the false positive rate is simply proportional to the false positive ($\sum^{+\infty}_{z\geq \lambda} \rho(z)- \text{true positive}$) in the input space. Thus, plotting the ROC curve is like plotting $1-r(z)$ and $r(z)$ scaled by $\rho(z)$ respectively. Comparing the equation of the (unnormalized) recall, this (unnormalized) false positive rate contains (almost if not all) the same information as the (unnormalized) recall in the input space.

\section{Language Classifier~\label{sup:sst}}
\begin{figure*}[ht]
  \centering
  \includegraphics[width=140mm]{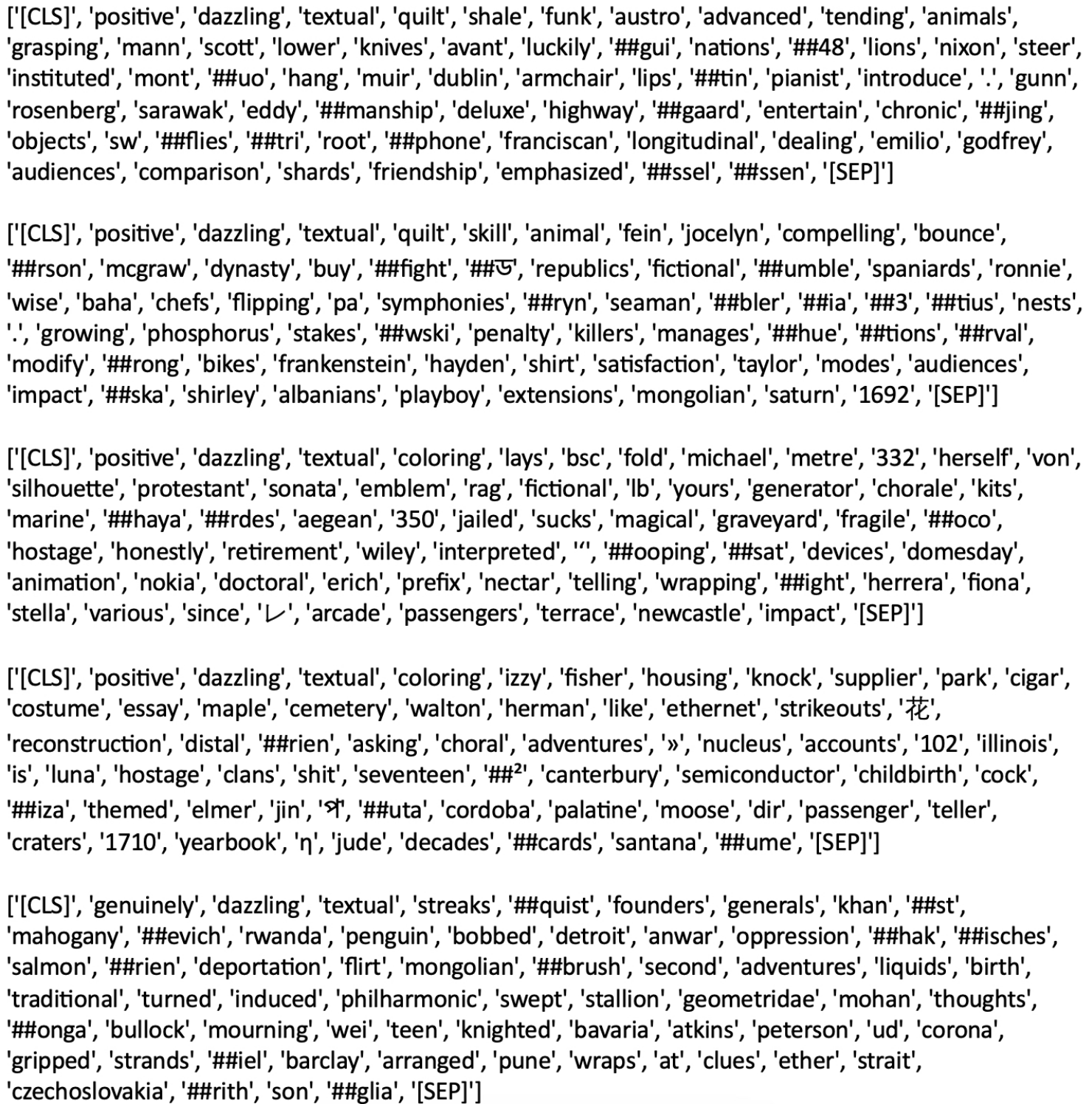}
  \caption{Sampled inputs of SST2 with sentence length 66.
  ~\label{fig:sst_long}}
\end{figure*}

For sentence length 66, some sampled inputs with logit equals 7 (positive sentiment) in Fig~\ref{fig:sst_long}.

\begin{figure*}[ht]
  \centering
  \includegraphics[width=140mm]{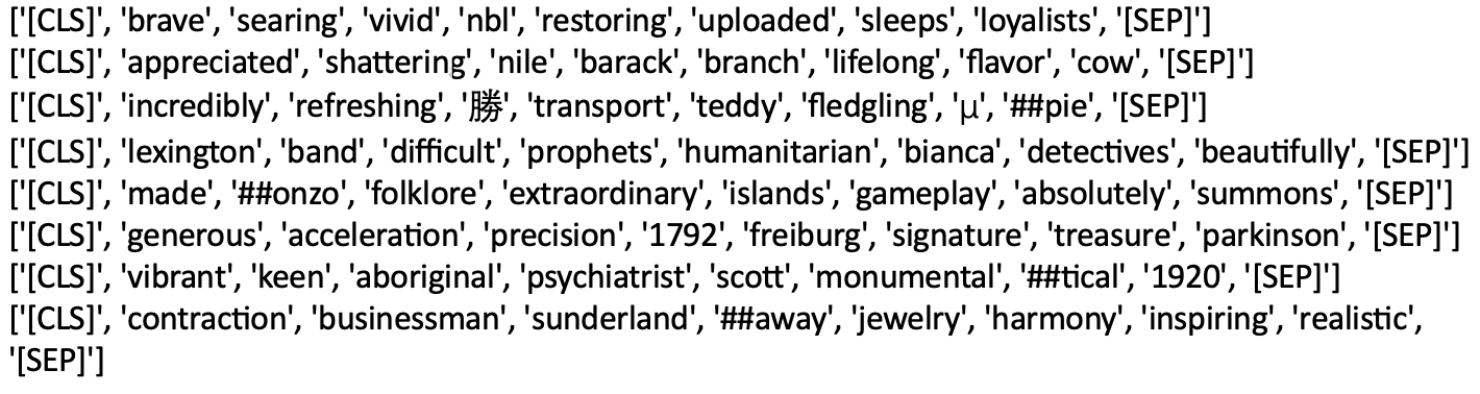}
  \caption{Sampled inputs of SST2 with sentence length 10.
  ~\label{fig:sst_short}}
\end{figure*}

For sentence length 10, some sampled inputs with logit equals 7 (positive sentiment) in Fig~\ref{fig:sst_short}.

\section{(Large) Language models~\label{app:llm}}
\begin{figure*}[ht]
\begin{center}
\subfigure[GPT2-small-25. Some information is masked as ``xxx.'']{\includegraphics[width=140mm]{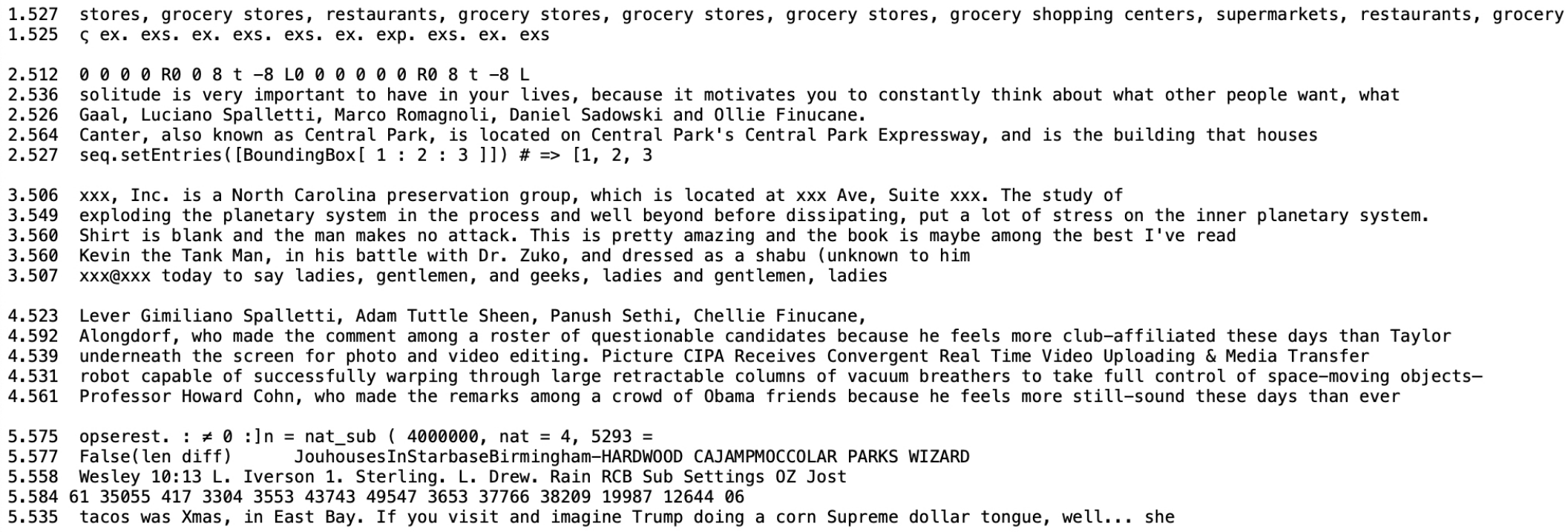}}
\subfigure[GPT2-small-100]{\includegraphics[width=140mm]{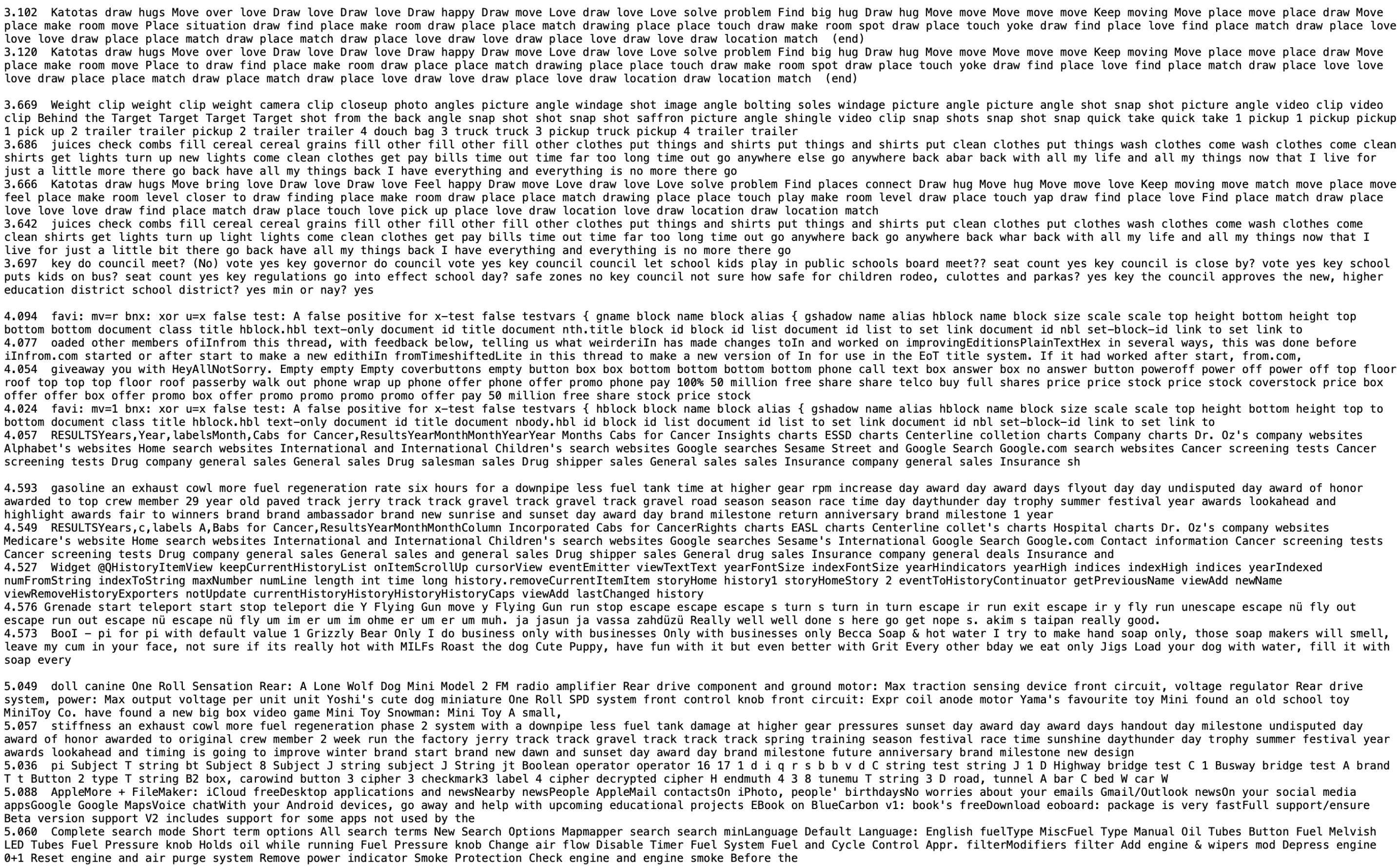}}
\subfigure[Llama2-25]{\includegraphics[width=140mm]{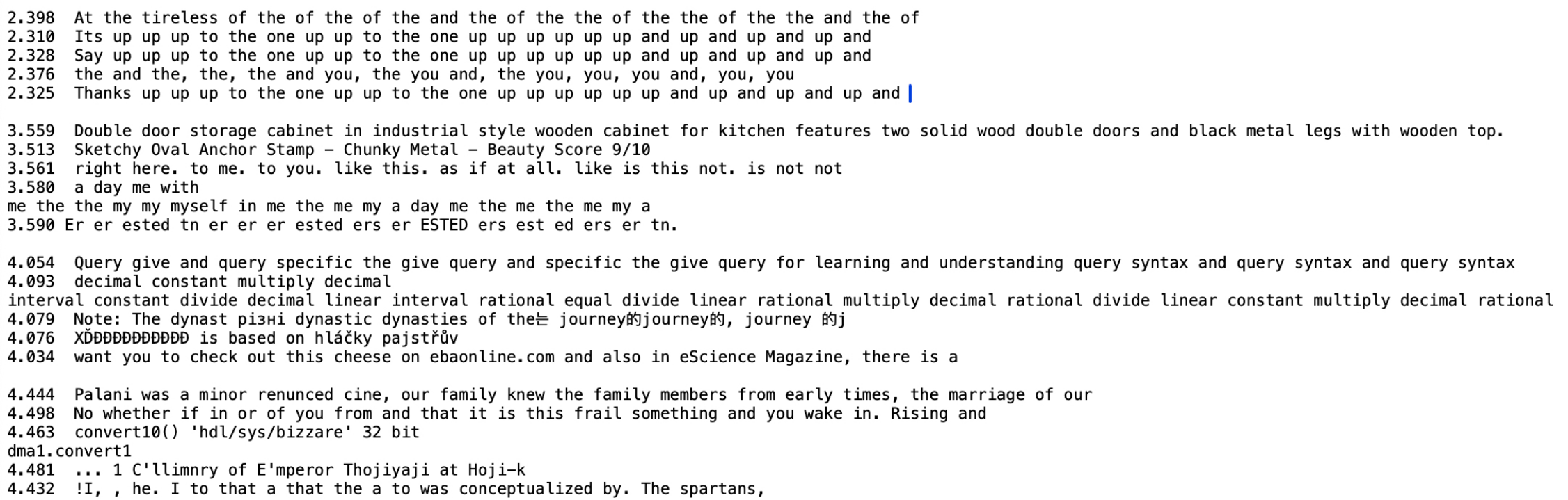}}
\subfigure[Llama1-25]{\includegraphics[width=140mm]{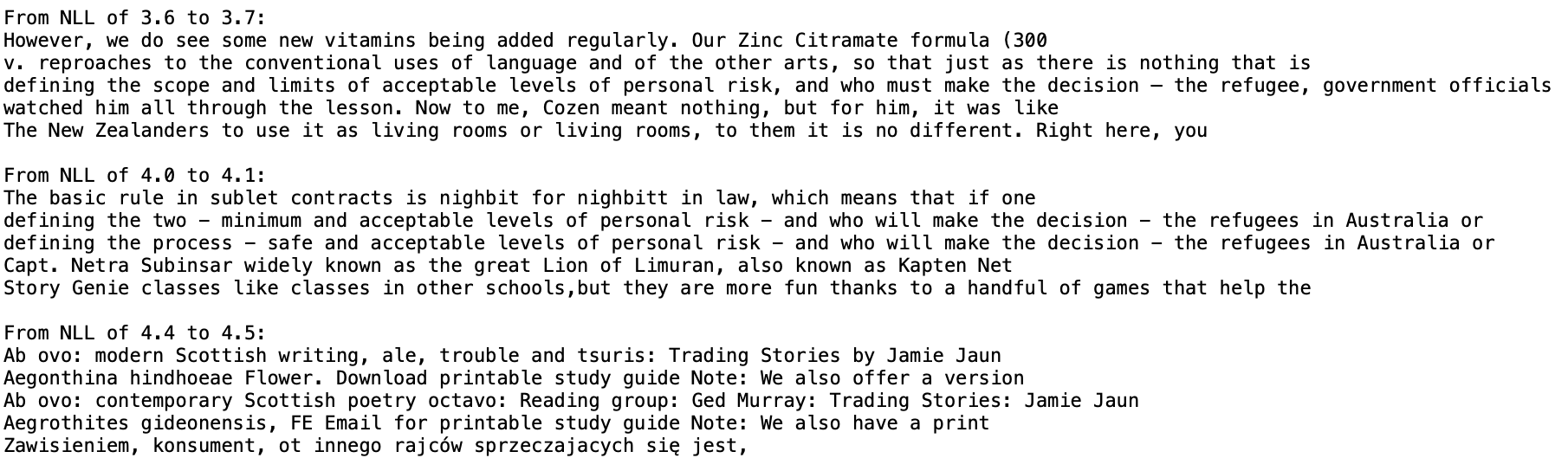}}
\caption{Some presentative inputs for different settings of language models. The NLLs are indicted as the numbers. ~\label{fig:llm}
}
\end{center}
\end{figure*}

Fig.~\ref{fig:llm} shows the sampled inputs for different settings. For GPT2-small-25, we scrutinize 30 bins with 30-300 non-duplicate inputs per bin. Every bin represents $\Delta$NLL=0.1. Here we show the sampled inputs for bin with NLL $[1.5, 1.6), [2.5,2.6), [3.5,3.6), [4.5,4.6), [5.5,5.6)$. When NLL is low, the model is simply repeating the meaningful phrases. When NLL is around 3.5, we observe the company address with an address which we cannot verify its correctness based on the internet. It also outputs the email address which we masked to protect privacy. When the NLL is high around 5.5, the sentences are much less understandable. 

For GPT2-small-100, we scrutinize 10 bins with 280-860 non-duplicate inputs per bin. Every bin represents $\Delta$NLL=0.1. For Llama2-25, we scrutinize 10 bins with 73-370 non-duplicate inputs per bin. Every bin represents $\Delta$NLL=0.1. When the NLL is low for both settings, the models seem to repeat a lot of phrases. When the NLL is high, the models seem to produce sentences that are hard to understand. 

\clearpage

\newpage

\end{document}